\documentclass[runningheads,a4paper]{llncs}

\usepackage{amssymb}
\usepackage{graphicx}

\usepackage{url}
\urldef{\mailsa}\path| {mina.rezaei, haojin.yang, christoph.meinel}@hpi.de|  
\newcommand{\keywords}[1]{\par\addvspace\baselineskip
\noindent\keywordname\enspace\ignorespaces#1}

\begin{document}

\mainmatter  

\title{Deep Learning for Medical Image Analysis}

\titlerunning{Deep Learning for Medical Image Analysis}

%
%
\author{Mina Rezaei, Haojin Yang, Christoph Meinel}
\authorrunning{Deep Learning for Medical Image Analysis}

\institute{Hasso Plattner Institute,\\
Prof.Dr.Helmert-Straße 2-3, 14482 Potsdam, Germany\\
\mailsa\\ }

%
%

\toctitle{Deep Learning for Medical Image Analysis}
\tocauthor{Authors' Instructions}
\maketitle

\begin{abstract}
This report describes my research activities in the Hasso Plattner Institute and summarizes my PhD plan and several novel, end-to-end trainable approches for analyze medical images using deep learning algorithm. In this report, as an example, we explore diffrent novel methods based on deep learning for brain abnormality detection, recognition and segmentation. This report prepared for doctoral consortium in AIME-2017 conferance.

\end{abstract}

\section{Introduction}

 During the past years deep learning has raised a huge attention by showing promising result in some state-of-the-art approaches such as speech recognition, handwritten character  recognition, image classification \cite{krizhevsky2012imagenet} \cite{He_2016_CVPR}, detection \cite{simonyan2014very} ,\cite{redmon2016you} and segmentation \cite{dai2016instance} \cite{long2015fully}. There are expectations that deep learning improve or create medical image analysis applications, such as computer aided diagnosis, image registration and multi-modal  image analysis, image segmentation and retrieval. There has been some application that using deep learning in medical application like cell tracking \cite{cciccek20163d} and organ cancer detection. Doctors use magnetic resonance images as effective tools to diagnosis the diseases. Computer aided medical diagnosis can perform fast and objective with high robustness and reliability to support the health system. The brain is particularly complex structure, and analysis of brain MR images is an important step for many diagnosis diseases and is excellent in early detection of cases of cerebral infarction, brain tumors, or infections. Automated brain lesion detection is an important clinical diagnostic task and very challenging because the lesions has different sizes, shapes, contrasts and locations. This paper reports our studies in three main sections of classification, localization and segmentation on brain diseases based on deep learning architecture.This paper introduces our contributions which is very fast and accurate and each of them depends the task can provides notable information about size, location and categories of lesions. We evaluated proposed methods on popular benchmark \cite{menze2015multimodal} and the outcome results compared with state-of-the art approches.

\section{Approach}
An advance medical application based on deep learning methods for diagnosis, detection, instance level semantic segmentation and even image synthesis from MRI to CT/X-ray is my goal. To do this I started with brain images, for lesion diagnosis, it consist of several steps. Classification, detection and segmentation are main steps. The flow chart in Figure 1 describes my PhD plan for this tasks based on deep leaning algorithms. Furthermore estimating generative models via an adversarial process \cite{goodfellow2016deep} could be next chapter of my PhD which now is very popular. My goal is to learn unsupervised learning and in first step by using GAN techniques, generate more data and use them as an augmented data for the training step. Image synthesis for generate diffrent modalities of MRI or for estimating CT images from MRI images by directly taking MR image patches as input and CT patches as output could be cover in this section. We would like to consider GAN approches for prediction tools during medicine treatment or chemo therapy for objective diseases like brain tumors, multiple sclerosis, liver tumor and etc. Practical application is in the last chapter. I would like to sum up my studies on an online platform and use data/model parallelism on GPU to make it more efficient and accessible for clinic and hospital or even radiologiest.

\begin{figure}
\includegraphics[width=0.9\textwidth]{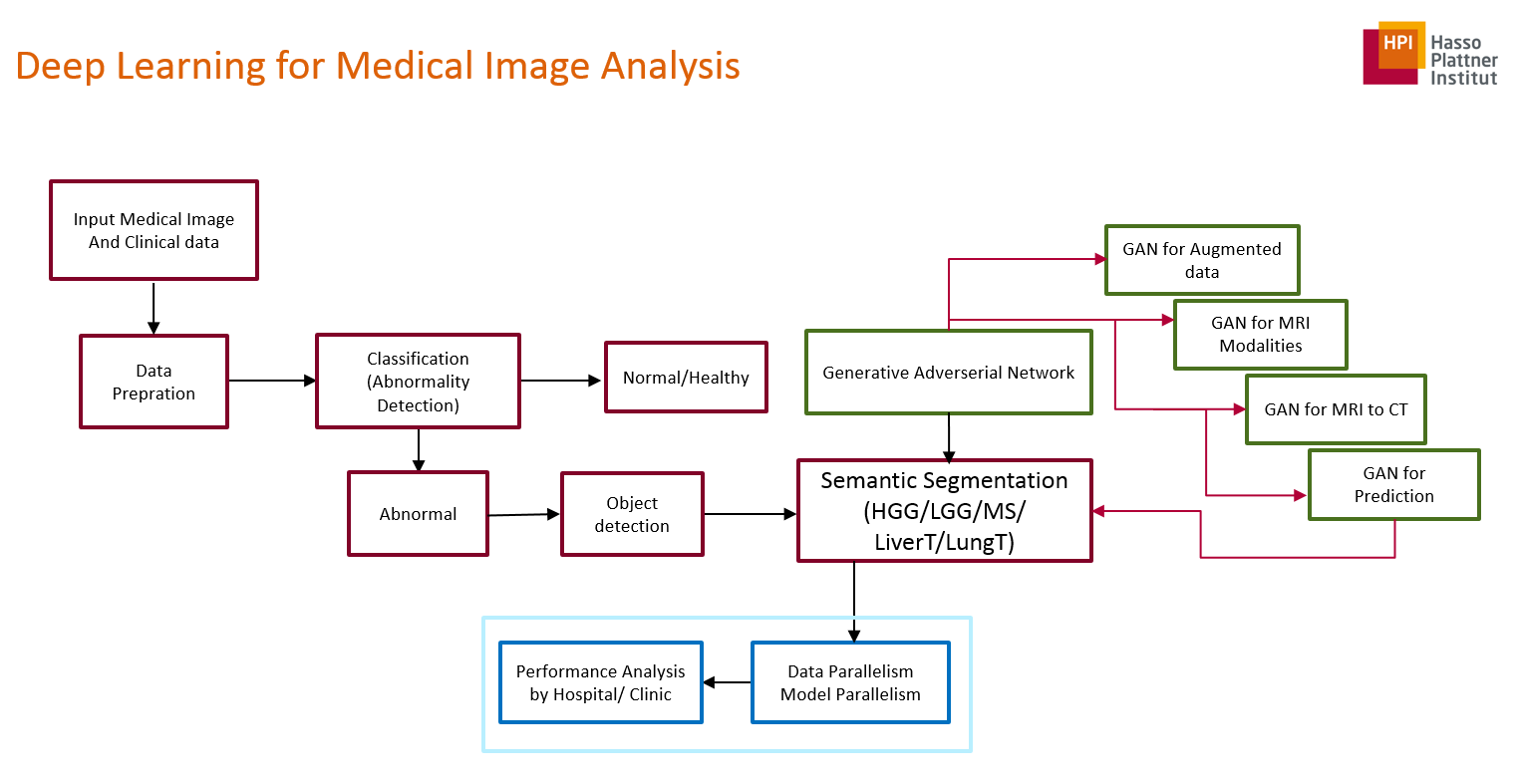}
\centering
\caption{Overall process for applying deep learning on medical image analysis}
\end{figure}
   
\subsection{Classification} For classification our current architecture is based on the network proposed by Krizhevsky et al \cite{krizhevsky2012imagenet}. The architecture is summarized in Figure 2. Input image for the network has three channels and we use axial, coronal and sagittal planes in a Volume-of-Interest (VOI) for each categories. In order to prevent over fitting in the next step we exploit data augmentation. I increase convolutional layer to seven to achieve better results.  I consider computation complexity and the time of process by training and testing using very efficient GPU implementation of the convolution operation. The network includes three pooling layers, average pooling after 3th convolution layer and two max pool layers consequently after 5th, 6th CNN layers that has effective impact on decreasing size of the feature map. At the end of 7th convolutional layer put three fully-connected layers which have 4096 individual neurons. By applying regularization after last fully connected layer to reduce chance of over fitting. In the final a 5-way SVM to classify 4 abnormalities and healthy brain images. Figure 2 presents current network for classification task. 

\begin{figure}
\includegraphics[width=0.9\textwidth]{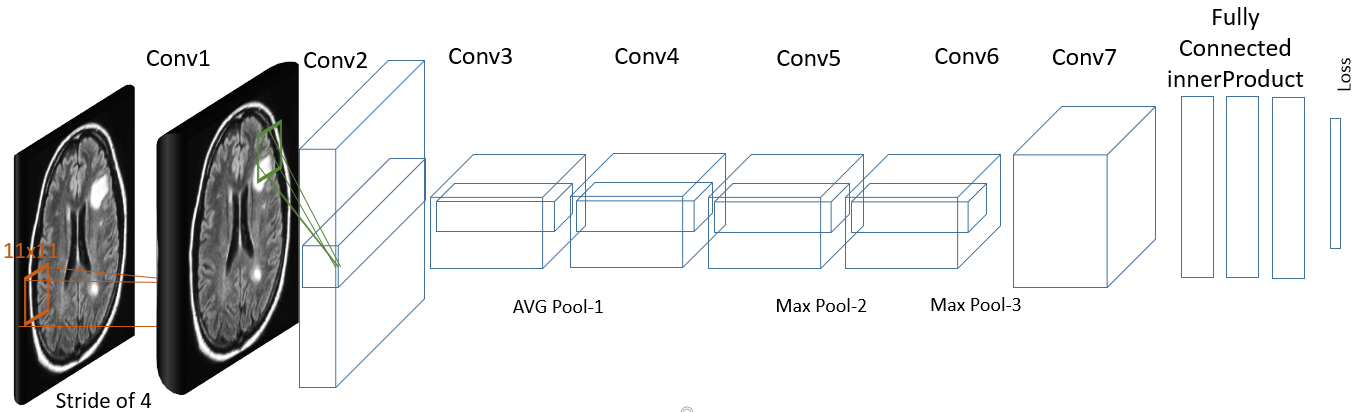}
\centering
\caption{A deep architecture designed to learn five different categories of MRI brain}
\end{figure}
\begin{table}[]
\centering
\caption{Classification result on five different MRI dataset by deep learning}
\label{my-label}
\begin{tabular}{|l|l|l|l|l|}
\hline
        ~ & Total MRI & accuracy & sensitivity & specificity \\
        Our propose method & 1500 & 95.07\% & 0.91 & 0.87  \\
\hline
\end{tabular}
\end{table}
\begin{table}[]
\centering
\caption{Comparing classification result with other approaches}
\label{my-label}
\begin{tabular}{|l|l|l|l|l|}
\hline
        ~ & Total MRI & Number of classes & accuracy  \\
        Our propose method & 1500 & 5 & 95.07\% \\
        Wavelet (DAUB-4) $+$ PCA $+$ SVM-RBF & 75 & 2 & 98.7\% \\  
        Wavelet (Haar) $+$ PCA $+$ KNN  & 1500 & 2 & 98.6\%  \\
\hline      
\end{tabular}
\end{table}
\begin{figure}
\includegraphics[width=0.9\textwidth]{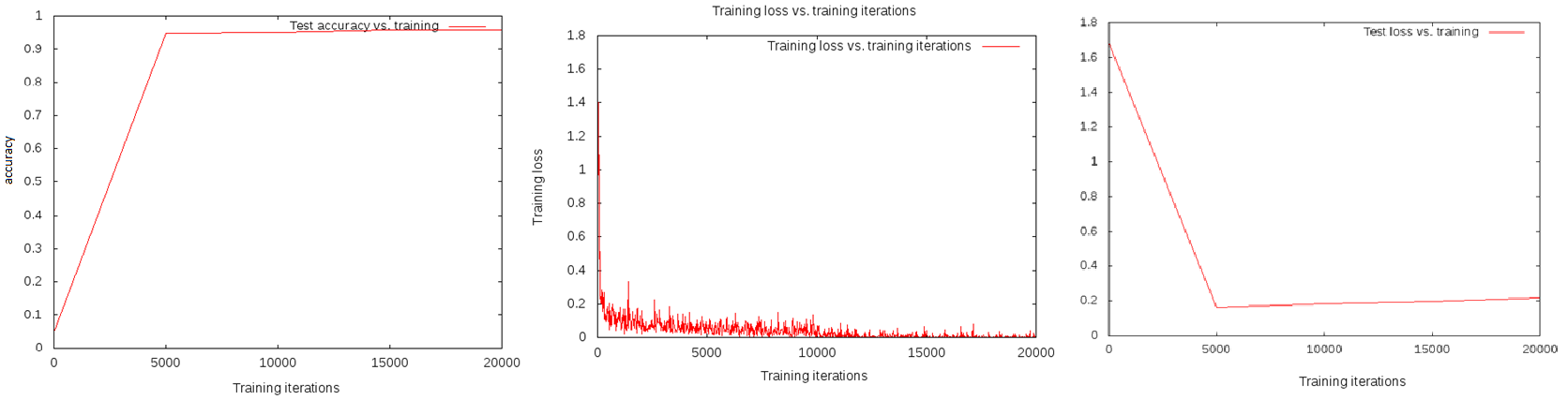}
\center
\caption{This charts describes accuracy,training and test loss according 20k iterations}
\end{figure}
\begin{figure}
\includegraphics[width=0.9\textwidth]{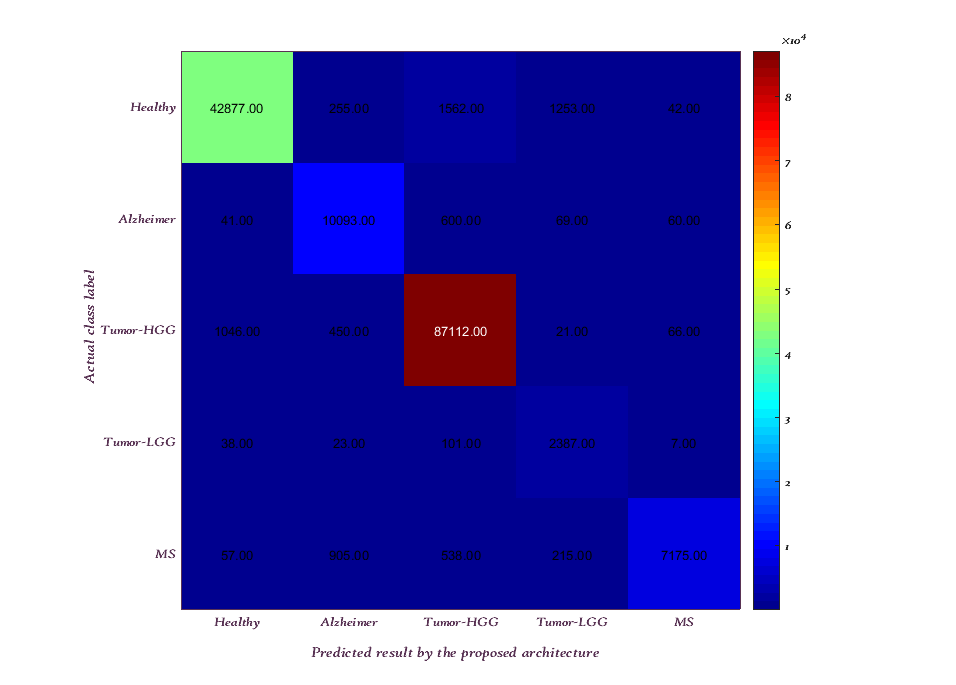}
\center
\caption{This charts describes accuracy,training and test loss according 20k iterations}
\end{figure}

\subsection{Detection and Localization}
 Unlike image classification, detection requires localizing (likely many) objects within an image. The method applied on brain magnetic resonance images (MRI) to recognize high and low grad glioma tumor, ischemic stroke and it can be extendable to detection and localization on other diseases like multiple sclerosis. Proposed method takes multi-modality 2D images in same volume of intresst for example for BraTs Benchmark this number 4. Same as classification task, the augm,ented data prepared in randomly flip horizontally and vertically, contrast changing and multi scaling . Additionaly this architecture is able to exploit local and global contextual features simultaneously and has achieved better results. The result shows by using multi modal images and combination of local features and global contextual features before fully connected the dice coefficient can improved up to 20\% in high and low grade glioma and up to 30\% in ischemic strike recognition.
The proposed method attacks the problem of brain abnormality detection by solving it slice by slice from the axial, coronal and sagittal view. Thus, the input x of the model corresponds to 2D image (slice), where each pixel is associated with multiple channels, each corresponding to a different image modalities. Then four different modalities of same patient in same slices passed to the two way convolutions. In the first way, there is only one convolution to exploit features from whole images. In the second way the patches of images enter to create a feature maps which in computer vision called it local features. The architecture in second path use Fast R-CNN \cite{girshick2015fast} as the base recognition network and fine-tuned VGG-16 \cite{simonyan2014very} for local feature extraction. As mentioned before l2-norm pooling in forward and backward path in both CPU and GPU implemented to generate another representation of features. In second path after conv5-3, is time to find region of interest (ROI) in the top of single layer of spatial pyramid pooling layers(SPP) \cite{he2014spatial} and then connected to fully connected layer at this time by concatenating this features with the features from whole images (from first network) in the fully connected. The multi task loss is calculated, which is based on both the bounding box repressors and Softmax classifier. This makes the layer even below the RoI pooling layer trainable. Figure 1 represent the deep architecture for brain abnormality detection. For evaluation by applied the model on BRATS-2015 dataset which contains 220 subjects with high grade and 54 subjects with low grade tumor. We have achieved dice 0.72, 0.89 sensitivity and 94.3 \% accuracy for whole tumor detection.  
\begin{figure}
\includegraphics[width=0.9\textwidth]{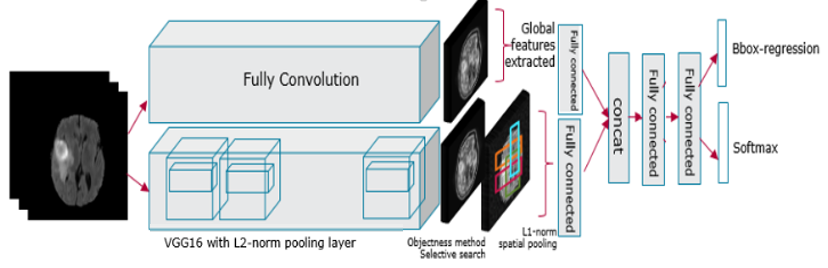}
\centering
\caption{Our proposed architecture by two path ways of convolution that can recognized and categorize brain lesion}
\end{figure}
\begin{figure}
\includegraphics[width=0.9\textwidth]{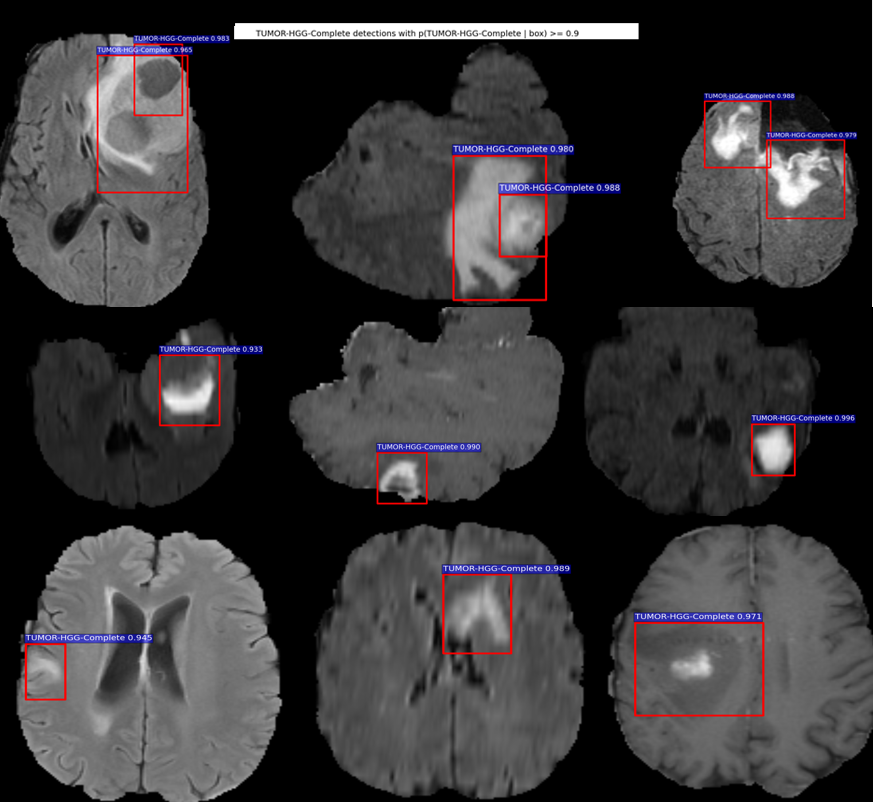}
\centering
\caption{Tumor high grade glioma detected by our CNN architecture}
\end{figure}
\begin{table}[]
\centering
\caption{shows the dice similarity coefficient (DSC) results on the BRATS2016 and ISLES 2016 by using different modalities. The table shows detection result can be improved 20\% in BRATS and 30\% in ISLES dataset.F/D coulmun means FLAIR modality in BraTS data set and DWI modality in ISLES dataset}
\begin{tabular}{|l|l|l|l|l|l|}
\hline
        T1 & T1c & T2 &F/D & Dice-BRATS16 &  Dice-ISLES  \\
        \hline
         x & -  & - & -  & 61.3 \% & 42\% \\
         - & x & - & - & 33.46 \% & 27\% \\
         - & - & x & - & 35.67 \% & 39.98\% \\
         - & - & - & x & 72.38 \% & 50.71\% \\
         - & x & x & x & 82.53 \% & 54.23\% \\
         x & x & - & x & 83.53 \% & 54.87\% \\
         x & - & x & x & 82.19 \% & 53.09\% \\
         x & x & x & - & 86.73 \% & 56.71\% \\
         x & x & x & x & 92.44 \% & 57.03\% \\ 
\hline
\end{tabular}
\end{table}
\begin{table}[]
\centering
\caption{Tumor high and low grade glioma detection and localization our proposed approach}
\begin{tabular}{|l|l|l|l|l|l|}
\hline
         ~ & FALIR & T1 & T2 &T1c& DWI \\
        \hline
        Tumor-HGG & 52990 & 52990 & 52990 & 52990& - \\
        Tumor-LGG & 35740& 35740 & 35740 &35740 & - \\
        ISLES &28500 & 28500 & - &28500 & 28500\\
\hline
\end{tabular}
\end{table}
\subsection{Semantic Segmentation} 
 We present very fast and accurate deep convolutional architecture for instance level segmentation which provides notable information about size, location and categories of the lesions as well. The proposed method takes 2D brain MR images from different modalities and some augmented data like randomly flip horizontally and vertically, contrast changing, multi scaling  and some generated data by multi conditional generative adversarial network which has designed by ourself as an input. This paper introduced two novel architecture, one for generating  augmented data and another for instance level segmentation of medical images.  The segmentation architecture is based on Faster r-cnn \cite{ren2015faster} which we make it faster by considering new strategy for anchors detection. Anchors detection, bounding box regression, mask level estimation and mask instance level segmentation is four stages for this task but it is end-to-end and single stage training by considering multi task loss function like \cite{dai2016instance}. We have achieved 89\% accuracy for non-enhancing tumor segmentation and table 5 describes the results in BRATS 2015/2016. 
\begin{figure}
\includegraphics[width=0.9\textwidth]{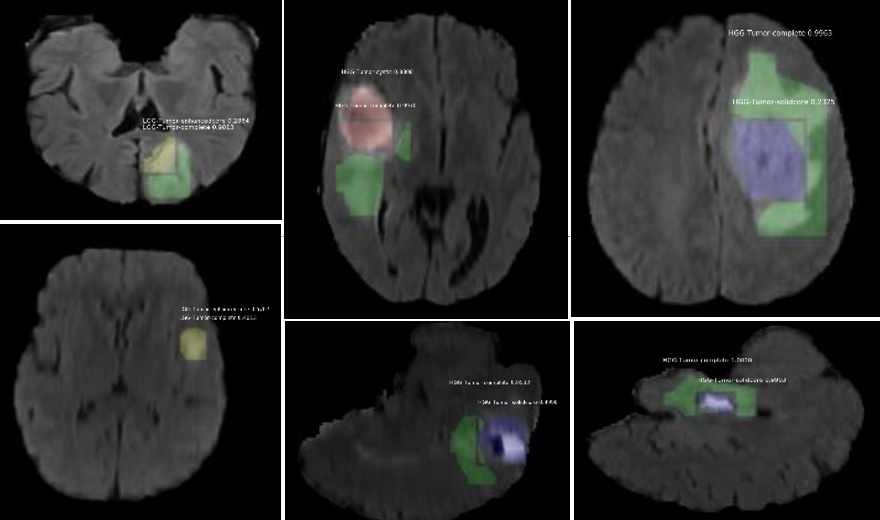}
\centering
\caption{The semantic/instance segmentation result by proposed method on Tumor high grade glioma}
\end{figure}

\begin{table}[]
\centering
\caption{the result of proposed method for segmentation  on Tumor high grade glioma }
\begin{tabular}{|l|l|l|l|l|}
\hline
         Accuracy & tumor-core & enhancing core & non-enhancing core & edema  \\
        \hline
        Bounding box-regression & 0.92831 & 0.9274  & 0.9407  & 0.9074 \\
        mask-estimation  & 0.88845 & 0.87132 & 0.92187 & 0.90862  \\
        instance-segmentation & 0.89562 & 0.84086 & 0.89375 & 0.89789 \\
\hline
\end{tabular}
\end{table}

\section{Data Description}
In this research I have used five different brain dataset to evaluate my proposed method.
\begin{enumerate}
   \item Healthy Brain Images: \footnote{http://brain-development.org/ixi-dataset/}This data has collected nearly 600 MR images from normal, healthy subjects. The MR image acquisition protocol for each subject includes:  
  \begin{itemize}
     \item T1, T2, T1-Contrast, PD-weighted images(Diffusion-weighted images in 15 directions)
   \end{itemize}
The data has been collected at three different hospitals in London and as part of IXI – Information extraction from Images project. The format of images is NIFTI (*.nii) and it is open access. Figure 1 column (a) shows healthy brain from IXI dataset in sagittal, coronal and axial section. 

  \item High and Low grade glioma(Tumor) \footnote{https://www.virtualskeleton.ch/BRATS/Start2015/}
  This data is from BRATS (Brain Tumor Segmentation) challenges in MICCAI conference 2015. The data prepared in two parts training and testing for high and low grade glioma tumor. All data sets have been aligned to the same anatomical template and interpolated to 1mm 3 voxel resolution. The training dataset contains about 300 high- and low- grade glioma cases. Each dataset has T1 (spin-lattice relaxation), T1 contrast-enhanced MRI, T2 (spin-spin relaxation), and FLAIR MRI volumes. In the test dataset 200 MR images without label but in the same format is available. Figure1 column (b, c) shows high and low grade glioma, both categories are .mha format.
  
 \item Alzheimer disease \footnote{http://www.medinfo.cs.ucy.ac.cy/index.php/downloads/datasets/}
  The Alzheimer data set downloaded from Open Access Series of Imaging Studies (OASIS). The dataset consists of a cross-sectional collection of 416 subjects aged 18 to 96.  For each subject, 3 or 4 individual T1-weighted MRI scans obtained in single scan sessions are included. The data format is *.hdr.
 
\item Multiple sclerosis\footnote{http://www.oasis-brains.org/}
  Multiple Sclerosis MR Images downloaded from ISBI conference 2008 (The MS Lesion Segmentation Challenges).This data set collected by e-Health lab of Cyprus University. Figure 6 column (e) shows multiple sclerosis images which training dataset consists of 18 multiple sclerosis as .nhdr format that ground truth (manual segmentation by expert) is available. 
\end{enumerate}

\begin{figure}
\includegraphics[width=0.7\textwidth]{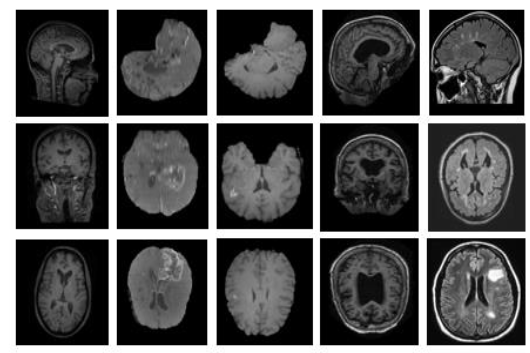}
\centering
\caption{We trained our architecture by five different categories of brain MRI. first column shows healthy brain in sagittal, coronal and axial section. second and third columns show tumor high and low grad glioma. In forth and fifth columns present some brain data on Alzheimer and multiple sclerosis}
\end{figure}

\section{Future Work}
 As first year PhD-student, highly intent to continue working on medical image analysis and first complete my studies on brain diseases. In machine learning and deep learning aspect I am currently doing 3D semantic segmentation on brain lesion. Data and model paralelism processingon GPU could be the next chapter of my thesis and my plan for second year. Generative adverserial network for Image generation and synthetisation is my plan for third year. Finaly after finding the optimum solution on brain diseases I would like to extend my result in other anatomy as well.

\bibliographystyle{splncs03}
\bibliography{typeinsts}

\end{document}